# Learning Bayesian Networks from Incomplete Databases

**Marco Ramoni**
Knowledge Media Institute
The Open University

**Paola Sebastiani**
Department of Actuarial Science and Statistics
City University

## Abstract

Bayesian approaches to learn the graphical structure of Bayesian Belief Networks (BBNs) from databases share the assumption that the database is complete, that is, no entry is reported as unknown. Attempts to relax this assumption involve the use of expensive iterative methods to discriminate among different structures. This paper introduces a deterministic method to learn the graphical structure of a BBN from a possibly incomplete database. Experimental evaluations show a significant robustness of this method and a remarkable independence of its execution time from the number of missing data.

## 1 INTRODUCTION

A Bayesian Belief Network (BBN) (Pearl, 1988) is a direct acyclic graph where nodes represent stochastic variables and arcs represent conditional dependencies among these variables. A conditional dependency links a *child* variable to a set of *parent* variables and is defined by the conditional distributions of the child variable given the configurations of its parent variables.

Although in their original concept BBNs were designed to rely on human experts to provide the graphical structure and assess the conditional probabilities, during the past few years an increasing number of efforts has been addressed toward the development of methods able to directly construct BBNs from databases. Early results in this quest were based on non Bayesian approaches (Sprites *et al.*, 1993), but a seminal paper by Cooper and Herskovitz (1992) gave rise to a stream of research within a Bayesian framework (Buntine, 1994; Heckerman *et al.*, 1995). Along this approach, the learning process involves two main tasks: the induction of the graphical model best fitting the database and the extraction of the conditional probabilities defining the dependencies in the graphical model.

Methods to perform the first task, known as *model selection*, typically involve two components: a search procedure to explore the space of possible graphical models and a scoring metric to assess the goodness-of-fit of a particular model. Current approaches exploit heuristics to reduce the search space and use the scoring metric to drive the search process. Although the task of extracting a BBN from a database in known to be NP-Hard for the general case (Chickering and Heckerman, 1994), under certain assumptions these methods are able to extract quite large BBNs from databases of thousands of cases. One of these assumptions is that the database is *complete*, that is, no entry in the database is reported as unknown.

The reason for this assumption is that a key step in the Bayesian learning process is the computation of the marginal likelihood of the database given a graphical model. This computation can be performed efficiently when the database is complete using exact Bayesian updating, but it becomes intractable when data are missing. Therefore, methods to approximate the marginal likelihood of the data have to be used. Current approaches (Chickering and Heckerman, 1996) exploit the EM algorithm (Dempster *et al.*, 1977) or Markov Chain Monte Carlo methods, such as Gibbs Sampling (Geman and Geman, 1984). The basic strategy underlying these methods is based on the *Missing Information Principle* (Little and Rubin, 1987): fill in the missing observations on the basis of the available information. EM performs this task by replacing the missing entries with the maximum likelihood estimates extracted from the available data and proceeds by iteratively estimating and replacing until stability is reached within a certain threshold. Gibbs Sampling generates a value for the missing data from some conditional distributions and provides a stochastic estimation of the posterior probabilities. Unfor-tunately, these methods are usually highly resource demanding, their convergence rate may be slow, and their execution time heavily depends on the number of missing data.

Ramoni and Sebastiani (1997b) introduced a deterministic method to estimate the conditional probabilities defining the dependencies in a BBN which does not rely on the Missing Information Principle. This method, called *Bound and Collapse* (BC), starts by *bounding* the set of possible estimates consistent with the available observations in the database and then *collapses* the resulting interval to a point via a convex combination of the extreme estimates with weights depending on the assumed pattern of missing data. The intuition behind BC is that the information available in the database induces a set of possible estimates and that the pattern of missing data can be used to select a single distribution within this set. The pattern of missing data may be either provided by an external source of information or may be estimated from the available information under the assumption that data are *missing at random*. Experimental evaluations (Ramoni and Sebastiani, 1997b) show clearly that the estimates provided by BC are very similar to those provided by Gibbs Sampling when data are missing at random, and they are more robust to departure from the true pattern of missing data. On the other hand, BC reduces the cost of estimating a conditional distribution to the cost of an exact Bayesian updating and a convex combination for each state of the distribution.

This paper describes how BC can be used to estimate the marginal likelihood of a database given a model thus extending the principle underlying BC from the task of learning the conditional probabilities to the task of extracting the graphical model of a BBN from an incomplete database. The reminder of this paper is structured as follows: Section 2 introduces the technical *background*, Section 3 describes the new *method*, Section 4 reports some results of a preliminary *experimental evaluation*, and Section 5 summarizes the relevant results.

## 2  BACKGROUND

A BBN is defined by a set of *variables* $\mathcal{X} = \{X_1, \ldots, X_I\}$ and a direct acyclic graph identifying a model $\mathcal{M}$ of conditional dependencies among these variables. A conditional dependency links a *child* variable $X_i$ to a set of *parent* variables $\Pi_i$, and is defined by the conditional distributions of the child variable given the configurations of its parent variables. We shall consider discrete variables only, and denote by $c_i$ the number of states of $X_i$, and $q_i$ the number of states of $\Pi_i$. The model $\mathcal{M}$ yields a factorization of the joint probability of a particular set of values of the variables in $\mathcal{X}$, say $x_k = \{x_{1k}, \ldots x_{Ik}\}$, as

$$p(\mathcal{X} = x_k | \mathcal{M}) = \prod_{i=1}^{I} p(X_i = x_{ik} | \Pi_i = \pi_{ij}, \mathcal{M}), \quad (1)$$

where $\pi_{ij}$ is the state of $\Pi_i$ in $x_k$. We will denote $X_i = x_{ik}$ by $x_{ik}$, and $\Pi_i = \pi_{ij}$ by $\pi_{ij}$.

Suppose we are given a database of $n$ cases $\mathcal{D} = \{x_1, \ldots, x_n\}$ from which we wish to select a model $\mathcal{M}$ of conditional dependencies among the variables in the database. We adopt a Bayesian approach, so that if $p(\mathcal{M})$ is our prior belief about a particular model $\mathcal{M}$, we can use the information in the database $\mathcal{D}$ to compute the posterior probability of $\mathcal{M}$ given the data:

$$p(\mathcal{M}|\mathcal{D}) = \frac{p(\mathcal{M}, \mathcal{D})}{p(\mathcal{D})},$$

and then we choose the model which has the highest posterior probability. When the comparison is between two rival models $\mathcal{M}_1$ and $\mathcal{M}_2$ with $p(\mathcal{M}_1) = p(\mathcal{M}_2)$, this is equivalent to choosing $\mathcal{M}_1$ if the Bayes factor:

$$\frac{p(\mathcal{D}|\mathcal{M}_1)}{p(\mathcal{D}|\mathcal{M}_2)} = \frac{p(\mathcal{M}_1, \mathcal{D})}{p(\mathcal{M}_2, \mathcal{D})},$$

is greater than one. It is well known (Cooper and Herskovitz, 1992), that $p(\mathcal{M}, \mathcal{D})$ can be easily computed if the conditional probabilities defining $\mathcal{M}$ are regarded as random variables $\theta_{ijk}$ whose prior distribution represents the observer's beliefs before seeing any data. The joint probability of a case $x_k$ can then be written in terms of the random vector $\theta = \{\theta_{ijk}\}$ as: $p(x_k|\theta) = \prod_{i=1}^{I} \theta_{ijk}$. This parameterization of the probabilities defining $\mathcal{M}$ allows us to write:

$$p(\mathcal{M}, \mathcal{D}) = p(\mathcal{M}) \int p(\theta|\mathcal{M}) p(\mathcal{D}|\theta) d\theta \quad (2)$$

where $p(\theta|\mathcal{M})$ is the prior density of $\theta$, and $p(\mathcal{D}|\theta)$ is the sampling model. A solution of (2) exists in closed form if: 1. *The database is complete;* 2. *The cases are independent, given the parameter vector $\theta$ associated to $\mathcal{M}$;* 3. *The prior distribution of the parameters is conjugate to the sampling model $p(\mathcal{D}|\theta)$;* 4. *The parameters are marginally independent.*

Let $n(x_{ik}|\pi_{ij})$, $i = 1, \ldots, I$, $j = 1, \ldots, q_i$, $k = 1, \ldots, c_i$, be the frequency of cases in the database with $x_{ik}|\pi_{ij}$, so that $n(\pi_{ij}) = \sum_{k=1}^{c_i} n(x_{ik}|\pi_{ij})$ is the frequency of cases with $\pi_{ij}$. Assumptions 1 and 2 lead to

$$p(\mathcal{D}|\theta) = \prod_{i=1}^{I} \prod_{j=1}^{q_i} \prod_{k=1}^{c_i} \theta_{ijk}^{n(x_{ik}|\pi_{ij})}.$$

A prior distribution on the parameters that satisfies 3 and 4 is a product of Dirichlet distributions. Thus,

if we denote by $\theta_{ij} = (\theta_{ij1}, ..., \theta_{ijc_i})$ the vector of parameters associated to the conditional distribution of $X_i|\pi_{ij}$, we have $\theta_{ij} \sim D(\alpha_{ij1}, ..., \alpha_{ijc_i})$. The prior hyper-parameters $\alpha_{ijk}$ can be regarded as frequencies of the imaginary cases needed to formulate the prior distribution. As a matter of fact, the marginal probability of $x_{ik}|\pi_{ij}$ is $\alpha_{ijk}/\alpha_{ij}$, and $\alpha_{ij} = \sum_{k=1}^{r_i} \alpha_{ijk}$ is the *prior precision* on $\theta_{ij}$. Under the assumptions 1 – 4, the posterior distribution of $\theta$ is still a product of Dirichlet distributions (Spiegelhalter and Lauritzen, 1990), and

$$\theta_{ij}|\mathcal{D} \sim D(\alpha_{ij1} + n(x_{i1}|\pi_{ij}), ..., \alpha_{ijc_i} + n(x_{ic_i}|\pi_{ij})).$$

Thus, the standard Bayesian estimate of $p(x_{ik}|\pi_{ij})$ is the *posterior expectation* of $\theta_{ijk}$:

$$E(\theta_{ijk}|\mathcal{D}) = \frac{\alpha_{ijk} + n(x_{ijk}|\pi_{ij})}{\alpha_{ij} + n(\pi_{ij})}, \tag{3}$$

and the *posterior precision* on $\theta_{ij}$ is $\alpha_{ij} + n(\pi_{ij})$. Furthermore, the integral (2) has the solution:

$$p(\mathcal{M}) \prod_{i=1}^{I} \prod_{j=1}^{q_i} \prod_{k=1}^{c_i} \frac{\Gamma(\alpha_{ij})\Gamma(\alpha_{ijk} + n(x_{ik}|\pi_{ij}))}{\Gamma(\alpha_{ij} + n(\pi_{ij}))\Gamma(\alpha_{ijk})}$$

and therefore

$$p(\mathcal{D}|\mathcal{M}) = \prod_{i=1}^{I} \prod_{j=1}^{q_i} \prod_{k=1}^{c_i} \frac{\Gamma(\alpha_{ij})\Gamma(\alpha_{ijk} + n(x_{ik}|\pi_{ij}))}{\Gamma(\alpha_{ij} + n(\pi_{ij}))\Gamma(\alpha_{ijk})} \tag{4}$$

is the *marginal likelihood* of $\mathcal{D}$ given $\mathcal{M}$. Note that $p(\mathcal{D}|\mathcal{M})$ depends on the updated hyper-parameters of $\theta_{ij}|\mathcal{D}$, and the posterior precision on $\theta_{ij}|\mathcal{D}$. The probability (4) is the base for the algorithm proposed by Cooper and Herskovitz (1992) to induce the model from a database. Suppose we have a partial order on the variables so that $X_i \prec X_j$ if $X_i$ cannot be parent of $X_j$. Let $\mathcal{P}_i$ be the set of current parents of $X_i$, thus $\mathcal{P}_i$ is the empty set if $X_i$ is a root node. Then the local contribution of a node $X_i$ and its parents $\Pi_i$ to the joint probability of $(\mathcal{M}, \mathcal{D})$ is measured by

$$g(X_i, \mathcal{P}_i) = \prod_{j=1}^{q_i} \prod_{k=1}^{c_i} \frac{\Gamma(\alpha_{ij})\Gamma(\alpha_{ijk} + n(x_{ik}|\pi_{ij}))}{\Gamma(\alpha_{ijk})\Gamma(\alpha_{ij} + n(\pi_{ij}))}. \tag{5}$$

The algorithm proceeds by adding a parent at a time and computing $g(X_i, \mathcal{P}_i)$. The set $\mathcal{P}_i$ is expanded to include the parent nodes that give the largest contribution to $g(X_i, \mathcal{P}_i)$, and stops if the probability does not increase any longer. This greedy search strategy has been shown to be extremely cost-effective when the number of variables is large. When the database is complete, (4) can be efficiently computed using the hyper-parameters $\alpha_{ijk} + n(x_{ik}|\pi_{ij})$ and the precision $\alpha_{ij} + n(\pi_{ij})$ of the posterior distribution of $\theta_{ij}$.

Suppose now that we are given the incomplete database $\mathcal{D}_i = \mathcal{D}_o \cup \mathcal{D}_m$, where $\mathcal{D}_m$ denotes the part of $\mathcal{D}_i$ with missing entries. The exact probability of $(\mathcal{M}, \mathcal{D}_i)$ is

$$p(\mathcal{M}, \mathcal{D}_i) = \sum_c p(\mathcal{M}, \mathcal{D}_i, \mathcal{D}_c) = \sum_c p(\mathcal{D}_i)p(\mathcal{M}, \mathcal{D}_c|\mathcal{D}_i)$$

where the sum is over all possible complete databases $\mathcal{D}_c$ consistent with the available data. Clearly, as the number of missing entries increases, the exact calculation of $p(\mathcal{M}, \mathcal{D}_i)$ is infeasible, and some approximation is needed.

## 3 METHOD

In this section, we will show that it is possible to approximate the hyper-parameters of the posterior distributions of $\theta_{ij}$, from which we derive an estimate of the marginal likelihood given in (4).

Let $\hat{p}_{ijk}$ be an estimate of the posterior expectation of $\theta_{ijk}$, and $\hat{\alpha}_{ij}$ be an estimate of the posterior precision of $\theta_{ij}$. Then the distribution $D(\hat{\alpha}_{ij}\hat{p}_{ij1}, ..., \hat{\alpha}_{ij}\hat{p}_{ijc_i})$ will have precision $\hat{\alpha}_{ij}$ and expectation $\hat{p}_{ijk}$, $k = 1, \cdots, c_i$. Thus a moment-matching approximation of the posterior distribution of $\theta_{ij}$ is:

$$\theta_{ij}|\mathcal{D} \sim D(\hat{\alpha}_{ij}\hat{p}_{ij1}, ..., \hat{\alpha}_{ij}\hat{p}_{ijc_i}). \tag{6}$$

From (6) we can then derive an estimate of (4):

$$\hat{p}(\mathcal{D}|\mathcal{M}) = \prod_{i=1}^{I} \prod_{j=1}^{q_i} \prod_{k=1}^{c_i} \frac{\Gamma(\alpha_{ij})\Gamma(\hat{\alpha}_{ij}\hat{p}(x_{ik}|\pi_{ij}))}{\Gamma(\alpha_{ijk})\Gamma(\hat{\alpha}_{ij})} \tag{7}$$

which can be also used to extend the algorithm in (Cooper and Herskovitz, 1992) to incomplete databases by estimating (5) as:

$$\hat{g}(X_i, \mathcal{P}_i) = \prod_{j=1}^{q_i} \prod_{k=1}^{c_i} \frac{\Gamma(\alpha_{ij})\Gamma(\hat{\alpha}_{ij}\hat{p}(x_{ik}|\pi_{ij}))}{\Gamma(\hat{\alpha}_{ij})\Gamma(\alpha_{ijk})}. \tag{8}$$

Clearly, the goodness of the approximation depends on the goodness of the estimates of $\hat{p}_{ijk}$ and $\hat{\alpha}_{ij}$. In the reminder of this section we will show how to use the BC method to estimate the posterior expectation of $\theta_{ijk}$, and the posterior precision of $\theta_{ij}$.

### 3.1 POSTERIOR EXPECTATION

Let $\mathcal{M}$ be a model of conditional dependencies, specifying for each $X_i$ the parent variable $\Pi_{ij}$. BC estimates the conditional probabilities defining the dependencies in $\mathcal{M}$ by first bounding the set of possible posterior distributions of $\theta_{ij}$ consistent with the database, and then collapsing the extreme distributions in one single Dirichlet using the assumed pattern of missing data.

| case | $X_1$ | $X_2$ | $X_3$ |
|------|-------|-------|-------|
| $x_1$ | 1 | 2 | 2 |
| $x_2$ | 2 | ? | 1 |
| $x_3$ | ? | 1 | 2 |
| $x_4$ | ? | ? | 1 |
| $x_5$ | 1 | ? | ? |

$$\Downarrow$$

$n^\bullet(x_{31}|(1,1)) = 2 \quad n^\bullet(x_{31}|(1,2)) = 2$

$n^\bullet(x_{31}|(2,1)) = 2 \quad n^\bullet(x_{31}|(2,2)) = 2$

$n^\bullet(x_{32}|(1,1)) = 2 \quad n^\bullet(x_{32}|(1,2)) = 1$

$n^\bullet(x_{32}|(2,1)) = 1 \quad n^\bullet(x_{32}|(2,2)) = 0$

Figure 1: Completions $n^\bullet(x_{3k}|x_1,x_2)$ consistent with the incomplete database.

Let $n^\bullet(x_{ik}|\pi_{ij})$ be the frequency of cases with $X_i = x_{ik}$, given the parent configuration $\pi_{ij}$, which have been obtained by completing the incomplete cases. A case may be incomplete because of either a missing observation in the parent configuration or a missing observation in the child variable. An example is given in Figure 1 for the model

$$X_1$$
$$\searrow \qquad X_i \quad \text{binary}, \quad i = 1,2,3.$$
$$X_2 \longrightarrow X_3$$

For each incomplete case, let $\phi_{ijk}$ be the probability of a completion:

$$\phi_{ijk} = p(x_{ik}|\pi_{ij}, X_i = ?). \tag{9}$$

When data are missing at random, and therefore $\mathcal{D}_o$ is a *representative sample* of the complete but unknown database $\mathcal{D}$, the probability of a completion can be estimated from $\mathcal{D}_o$ as

$$\hat{\phi}_{ijk} = \frac{\alpha_{ijk} + n(x_{ik}|\pi_{ij})}{\alpha_{ij} + \sum_h n(x_{ih}|\pi_{ij})}.$$

In this case, the BC estimate $\hat{p}(x_{ik}|\pi_{ij}, \mathcal{D}_i, \phi_{ijk})$ of $E(\theta_{ik}|\mathcal{D}_i)$ becomes:

$$\sum_{l \neq k} \hat{\phi}_{ijl} p_{l\bullet}(x_{ik}|\pi_{ij}, \mathcal{D}_i) + \hat{\phi}_{ijk} p^\bullet(x_{ik}|\pi_{ij}, \mathcal{D}_i) \tag{10}$$

where

$$p^\bullet(x_{ik}|\pi_{ij}, \mathcal{D}_i) = \frac{\alpha_{ijk} + n(x_{ik}|\pi_{ij}) + n^\bullet(x_{ik}|\pi_{ij})}{\alpha_{ij} + \sum_h n(x_{ih}|\pi_{ij}) + n^\bullet(x_{ik}|\pi_{ij})}$$

$$p_{l\bullet}(x_{ik}|\pi_{ij}, \mathcal{D}_i) = \frac{\alpha_{ijk} + n(x_{ik}|\pi_{ij})}{\alpha_{ij} + \sum_h n(x_{ih}|\pi_{ij}) + n^\bullet(x_{il}|\pi_{ij})}.$$

The value $p^\bullet(x_{ik}|\pi_{ij}, \mathcal{D}_i)$ is the upper bound of $p(x_{ik}|\pi_{ij}, \mathcal{D}_i)$, which is achieved when all incomplete cases in the database which could be completed as $x_{ik}|\pi_{ij}$ are assigned to $x_{ik}|\pi_{ij}$, and the other incomplete cases are assigned to $x_{ih}|\pi_{il}$, any $h$, and $l \neq j$. Thus, each maximum probability $p^\bullet(x_{ik}|\pi_{ij}, \mathcal{D}_i)$ is obtained from a Dirichlet distribution

$$D_k(\alpha_{ij1} + n(x_{i1}|\pi_{ij}), \ldots, \alpha_{ijk} + n(x_{ik}|\pi_{ij}) + $$
$$n^\bullet(x_{ik}|\pi_{ij}), \ldots, \alpha_{ijc_i} + n(x_{ic_i}|\pi_{ij}))$$

which identifies a unique probability $p_{k\bullet}(x_{il}|\pi_{ij}, \mathcal{D}_i)$ for the other states of the variable $X_i$ given $\pi_{ij}$ from which $p_{l\bullet}(x_{ik}|\pi_{ij}, \mathcal{D}_i)$ is obtained. The estimates $\hat{p}(x_{ik}|\pi_{ij}, \mathcal{D}_i, \phi_{ijk})$, $k = 1, \ldots, c_i$, so found define a probability distribution since $\sum_{k=1}^{c_i} \hat{p}(x_{ik}|\pi_{ij}, \mathcal{D}_i, \phi_{ijk}) = 1$.

As the number of missing entries in $\mathcal{D}_i$ decreases, $p^\bullet(x_{ik}|\pi_{ij}, \mathcal{D}_i)$ and $p_{l\bullet}(x_{ik}|\pi_{ij}, \mathcal{D}_i)$ approach $(\alpha_{ijk} + n(x_{ik}|\pi_{ij}))/(\alpha_{ij} + n(\pi_{ij}))$ so that, when the database is complete, (10) returns the exact estimate $E(\theta_{ijk}|\mathcal{D}_i)$. As the number of missing entries increases then both $\hat{\phi}_{ijk}$ and the estimate (10) approach the prior probability $\alpha_{ijk}/\alpha_{ij}$, so that the estimation method is coherent and no updating is performed when data are totally missing.

If $n^\bullet(x_{ik}|\pi_{ij}) = n^\bullet_{ij}$, as for instance when data are missing only on the child variable, (10) simplifies to

$$\frac{\alpha_{ijk} + n(x_{ik}|\pi_{ij}) + \hat{\phi}_{ijk} n^\bullet_{ij}}{\alpha_{ij} + \sum_h n(x_{ih}|\pi_{ij}) + n^\bullet_{ij}}, \tag{11}$$

which is a consistent estimate of the *expected posterior expectation*

$$\frac{x_{ijk} + n(x_{ik}|\pi_{ij}) + n^\bullet_{ij} \phi_{ijk}}{\alpha_{ij} + \sum_h n(x_{ih}|\pi_{ij}) + n^\bullet_{ij}}.$$

If $\alpha_{ijk} = 0$, then (11) is the classical maximum likelihood estimate of $\theta_{ijk}$ (Little and Rubin, 1987). Experimental comparisons (Ramoni and Sebastiani, 1997b) have shown that, when data are missing at random, the estimates computed by the BC method are very close to those obtained by Gibbs Sampling, and are more robust to departures from the true pattern of missing data.

Although BC is able to incorporate the assumption that data are missing at random, in the general case it is not limited to it, since the parameters $\phi_{ijk}$ may be used to encode any pattern of missing data. For instance, when no information on the mechanism generating the missing data is available and therefore any pattern is equally likely, then $\phi_{ijk} = 1/c_i$. Furthermore, BC provides a new measure of the information available in the database when we consider that the extreme probabilities $p_{l\bullet}(x_{ik}|\pi_{ij}, \mathcal{D}_i)$, $l = 1, \cdots, c_i$ lead to a lower bound of $p(x_{ik}|\pi_{ij}, \mathcal{D}_i)$ — that is, $p_\bullet(x_{ik}|\pi_{ij}, \mathcal{D}_i) = \min_l \{p_{l\bullet}(x_{ik}|\pi_{ij}, \mathcal{D}_i)\}$ — and therefore the interval $[p_\bullet(x_{ik}|\pi_{ij}, \mathcal{D}_i), p^\bullet(x_{ik}|\pi_{ij}, \mathcal{D}_i)]$ contains all posterior estimates of $\theta_{ijk}$ that would be obtained from the possible completions of the database,

|            | Generating Structure | Variables | Cases |
|------------|----------------------|-----------|-------|
| $\mathcal{M}_1$ | $X_1 \rightarrow X_2 \rightarrow X_3$ | $X_1(2)X_2(2)$ $X_3(2)$ | 1000 |
| $\mathcal{M}_2$ | $X_1 \rightarrow X_2 \rightarrow X_3$ | $X_1(2)X_2(2)$ $X_3(3)$ | 1000 |
| $\mathcal{M}_3$ | $X_5 \leftarrow X_3 \quad X_4$ $\nwarrow \quad \nearrow$ $X_1 \rightarrow X_2$ | $X_1(2)X_2(2)$ $X_3(3)$ $X_4(2)X_5(2)$ | 5000 |
| $\mathcal{M}_4$ | $X_5 \leftarrow X_3 \quad X_4$ $\nwarrow \quad \nearrow$ $X_1 \rightarrow X_2$ | $X_1(3)X_2(3)$ $X_3(3)$ $X_4(3)X_4(2)$ | 10000 |

Table 1: Generating structures used in the experimental evaluations. The number next to each variable reports the number of states.

thus providing a measure of the quality of information conveyed by $\mathcal{D}_i$ about $\theta_{ijk}$ (Ramoni and Sebastiani, 1997a).

## 3.2   POSTERIOR PRECISION

The value in (10) is an estimate of $E(\theta_{ijk}|\mathcal{D}_i)$. We now derive an estimate of the posterior precision of $\theta_{ij}$. Suppose we have $n(\pi_{ij})$ cases completely observed on $\pi_{ij}$, so that $n - \sum_j n(\pi_{ij})$ is the number of cases partially observed on the parent variable $\Pi_i$. Let $\theta_i = (\theta_{i1}, ..., \theta_{iq_i})$ be the parameters associated to the joint probability distribution of $\Pi_i$, and let $D(\beta_{i1}, ..., \beta_{iq_i})$ be the prior distribution, so that $\beta_i = \sum_j \beta_{ij}$ is the prior precision. If we knew the probability distribution of $\pi_{ij}$ we could distribute the incomplete cases across the states of $\Pi_i$, so that the expected precision of the posterior distribution of $\theta_{ij}$ would be $\alpha_{ij} + n(\pi_{ij}) + p(\pi_{ij})(n - \sum_j n(\pi_{ij}))$. Thus if $\hat{p}(\pi_{ij}|\mathcal{D}_i)$ is an estimate of $p(\pi_{ij})$, an estimate of the posterior precision is

$$\hat{\alpha}_{ij} = \alpha_{ij} + n(\pi_{ij}) + \hat{p}(\pi_{ij}|\mathcal{D}_i)(n - \sum_j n(\pi_{ij})). \quad (12)$$

Clearly, $\hat{\alpha}_{ij}$ is the exact posterior precision when the database is complete and, as the number of missing entries increases, the accuracy of $\hat{\alpha}_{ij}$ heavily depends on $\hat{p}(\pi_{ij}|\mathcal{D}_i)$. We can apply the BC method to obtain the estimate $\hat{p}(\pi_{ij}|\mathcal{D}_i)$. When data are missing at random, the estimate of $\phi_{ij} = p(\Pi_i = \pi_{ij}|\Pi =?)$, $j = 1, ..., q_i$, is

$$\hat{\phi}_{ij} = \frac{\beta_{ij} + n(\pi_{ij})}{\beta_i + \sum_h n(\pi_{ih})}.$$

We can then apply (10) to obtain

$$\hat{p}(\pi_{ij}) = \sum_{l \neq j=1}^{q_i} \hat{\phi}_{il} p_{l\bullet}(\pi_{ij}|\mathcal{D}_i) + \hat{\phi}_{ij} p^\bullet(\pi_{ij}|\mathcal{D}_i)$$

| %   | Induced Model | $\hat{l}(\mathcal{D}_i|\mathcal{M})$ | Time |
|-----|---------------|--------------------------------------|------|
| 100 | $X_1 \longrightarrow X_2 \longrightarrow X_3$ | 1437 | 12 |
| 80  | $X_2 \longrightarrow X_3$ $\uparrow \quad \nearrow$ $X_1$ | 1426 | 13 |
| 60  | $X_1 \quad X_2 \longrightarrow X_3$ | 1446 | 11 |
| 40  | $X_1 \longrightarrow X_2 \longrightarrow X_3$ | 1447 | 12 |
| 20  | $X_1 \longrightarrow X_2 \longrightarrow X_3$ | 1414 | 12 |

Table 2: Models induced from the database generated from $\mathcal{M}_1$ for different percentages of available entries.

where

$$p^\bullet(\pi_{ij}|\mathcal{D}_i) = \frac{\beta_{ij} + n(\pi_{ij}) + n^\bullet(\pi_{ij})}{\beta_i + \sum_h n(\pi_{ih}) + n^\bullet(\pi_{ij})}$$

$$p_{l\bullet}(\pi_{ij}|\mathcal{D}_i) = \frac{\beta_{ij} + n(\pi_{ij})}{\beta_i + \sum_h n(\pi_{ih}) + n^\bullet(\pi_{il})},$$

with $n^\bullet(\pi_{ij})$ denoting the number of possible completions of the incomplete cases on $\pi_{ij}$. As the number of missing entries increases, the estimate $\hat{\alpha}_{ij}$ tends to $\alpha_{ij} + (\beta_{ij}/\beta_i)n$ so that the cases are distributed according to the prior belief about the parameters defining the BBN.

## 4   EXPERIMENTAL EVALUATION

The aim of the experiments described in this Section is to evaluate the accuracy of the estimate (7) as the number of missing entries in the database increases.

### 4.1   MATERIALS AND METHODS

We considered four different models described in Table 1. From each of these models we generated a random sample of $n$ cases, and applied the algorithm for the induction of the model from the data, using an initial order which was consistent with the generating structure, and assuming uniform prior distributions on the parameters. We then iteratively deleted 20% of the sample at random, until the database was empty. On each incomplete database we run our system to induce the model from the data. The algorithm takes as input a database together with a partial order on the variables occurring in it, and returns a BBN. The induction of the graphical model uses a greedy search strategy and replaces the measure (5) with the BC estimate (8). Once the graphical model has been chosen, the conditional probabilities are estimated using the BC method. This method was implemented in Common Lisp and the experiments were performed on a Macintosh 7500/100.

| % | $X_1 = 1$ | $X_2 = 1$ | $X_3 = 1$ |
|---|---|---|---|
| 100 | 0.11 | 0.78 | 0.56 |
| 80 | 0.11 | 0.78 | 0.57 |
| 60 | 0.12 | 0.79 | 0.56 |
| 40 | 0.11 | 0.79 | 0.57 |
| 20 | 0.10 | 0.79 | 0.60 |

Table 3: Marginal probabilities induced for the structure $\mathcal{M}_1$ for different percentages of available entries.

| | 100 | 80 | 60 | 40 | 20 |
|---|---|---|---|---|---|
| $g(X_1)$ | 356 | 353 | 367 | 350 | 324 |
| $g(X_2)$ | 531 | 526 | 519 | 512 | 506 |
| $g(X_3)$ | 690 | 692 | 689 | 689 | 678 |
| $g(X_2, X_1)$ | 519 | 512 | 520 | 511 | 483 |
| $g(X_3, X_1)$ | 691 | 692 | 692 | 684 | 667 |
| $g(X_3, X_2)$ | 562 | 560 | 560 | 586 | 607 |
| $g(X_3, (X_1, X_2))$ | 564 | 554 | 566 | 593 | 609 |

Table 4: Estimate of $-\log g(X_i, \Pi)$ for different percentages of available entries in the database generated from $\mathcal{M}_1$.

## 4.2 RESULTS AND DISCUSSION

Tables 2 and 6 show the models induced from the databases generated from the two models $\mathcal{M}_1$ and $\mathcal{M}_2$, the estimates of $-\log p(\mathcal{D}_i|\mathcal{M})$ for different percentages of available entries, and the total run time, in seconds, taken to extract the graphical model and estimate the parameters of the BBN. Tables report $-\log \hat{p}$ as $\hat{l}$. The marginal probabilities are displayed in Tables 3 and 7. The initial order on the variables was in both cases $X_3 \prec X_2 \prec X_1$.

The models learned from the database generated from $\mathcal{M}_1$ are the correct ones when 40% and 20% of the entries in the database are available, and coherently the model of independence is induced from the empty database. With 60% and 80% of the entries, the induced models differ from the generating structure in one link. Run times show a remarkable independence from the percentage of missing data in the database.

Table 4 gives the estimates $-\log \hat{g}(X_i, \Pi_{ij})$ computed in each step of the algorithm. When 80% of the entries is available, $-\log \hat{g}(X_3, (X_1, X_2)) = 554$ and $-\log \hat{g}(X_3, X_2) = 560$, so that the model induced from the incomplete database is $\exp(-554 + 560) = 403.4$ times more likely than the generating structure, if we assume that the prior distribution on the eight possible models consistent with the order $X_3 \prec X_2 \prec X_1$ is uniform. The strong evidence against the model used to generate the database can be due to the fact that $p(X_3 = 1|X_2 = 2) = 0.1$ and $p(X_2 = 2) = 0.77$ in the generating structure. In the complete database $n(X_3 = 1|X_2 = 2) = 22$ which becomes 11 when 20% of entries are deleted, so that the small number of entries may cause the imprecision of the estimate $-\log \hat{p}(\mathcal{D}_i|\mathcal{M})$. The conditional probabilities estimated for the model selected are $p(X_3 = 1|X_1 = 1, X_2 = 1) = 0.77$ and $p(X_3 = 1|X_1 = 2, X_2 = 1) = 0.70$, $p(X_3 = 1|X_1 = 1, X_2 = 2) = 0.12$ and $p(X_3 = 1|X_1 = 2, X_2 = 2) = 0.11$, so that the estimate of the marginal probability of $X_3 = 1$ differs from the estimate obtained from the complete database by 1%. When 60% of the entries are available $-\log \hat{g}(X_2) = 519$ and $-\log \hat{g}(X_2, X_1) = 520$ so that the model induced from the data is only 2.7 times more likely than the generating structure. Again the marginal probabilities computed from the induced network are very similar to the marginal probabilities found in the model induced from the complete database: thus the choice of a slightly different model has little effect on the predicting power of the network.

| | 100 | 80 | 60 | 40 | 20 |
|---|---|---|---|---|---|
| $X_1 \quad X_3 \quad X_2$ | 1577 | 1570 | 1575 | 1552 | 1508 |
| $X_1 \to X_3 \quad X_2$ | 1579 | 1571 | 1578 | 1546 | 1491 |
| $X_1 \quad X_2 \to X_3$ | 1450 | 1439 | 1446 | 1448 | 1437 |
| $X_1 \to X_3 \leftarrow X_2$ | 1452 | 1432 | 1452 | 1455 | 1438 |
| $X_1 \to X_2 \quad X_3$ | 1565 | 1557 | 1576 | 1551 | 1485 |
| $X_2 \leftarrow X_1 \to X_2$ | 1566 | 1557 | 1578 | 1545 | 1469 |
| $X_1 \to X_2 \to X_3$ | 1437 | 1425 | 1447 | 1447 | 1414 |
| $X_1 \to X_2$ <br> $\downarrow \swarrow$ <br> $X_3$ | 1439 | 1419 | 1452 | 1454 | 1416 |

Table 5: $-\log \hat{p}(\mathcal{D}_i|\mathcal{M})$ for all possible models consistent with $X_3 \prec X_2 \prec X_1$, for different percentages of available entries generated from $\mathcal{M}_1$.

| % | Induced Model | $\hat{l}(\mathcal{D}_i|\mathcal{M})$ | Time |
|---|---|---|---|
| 100 | $X_1 \longrightarrow X_2 \longrightarrow X_3$ | 1869 | 12 |
| 80 | $X_2 \longrightarrow X_3$ <br> $\uparrow \quad \nearrow$ <br> $X_1$ | 1855 | 13 |
| 60 | $X_1 \longrightarrow X_2 \longrightarrow X_3$ | 1865 | 11 |
| 40 | $X_2 \longrightarrow X_3$ <br> $\uparrow \quad \nearrow$ <br> $X_1$ | 1825 | 12 |
| 20 | $X_1 \longrightarrow X_2 \longrightarrow X_3$ | 1770 | 12 |

Table 6: Models induced from the database generated from $\mathcal{M}_2$ for different percentages of available entries.

Table 5 gives the estimate $-\log \hat{p}(\mathcal{D}_i|\mathcal{M})$ for the eight possible models consistent with the initial ordering of the variables. These estimates can be computed from

| %   | $X_1 = 1$ | $X_2 = 1$ | $X_3 = 1$ | $X_3 = 2$ |
|-----|-----------|-----------|-----------|-----------|
| 100 | 0.11      | 0.78      | 0.25      | 0.30      |
| 80  | 0.12      | 0.78      | 0.23      | 0.30      |
| 60  | 0.12      | 0.79      | 0.23      | 0.29      |
| 40  | 0.12      | 0.79      | 0.23      | 0.28      |
| 20  | 0.10      | 0.81      | 0.19      | 0.30      |

Table 7: Marginal probabilities in the networks induced from the database generated from $\mathcal{M}_2$ for different percentages of available entries.

the values in Table 4 by adding relevant terms. The estimates are very accurate until 40% of the entries are retained. When only 20% of the entries is available, the error of the estimate increases, but nonetheless the model induced from the database is equal to the generating structure. If we assume that the set of possible models is limited to the eight models consistent with the order $X_3 \prec X_2 \prec X_1$, and that they are a priori equally likely, then from the values in Table 5 we can compute the marginal probability of $\mathcal{D}$ and of the four incomplete databases $\mathcal{D}_i$ from which we can compute the posterior probabilities of all possible models. The posterior probability of the model induced from the database with 80% of the entries is 0.9987, against a probability 0.0012 for the generating structure. The other models have posterior probabilities near 0. With 60% of the entries, the posterior probability of the induced model is 0.6699, against 0.3258 for the generating structure.

Similar results are found for the models induced from the database generated from $\mathcal{M}_2$. The models induced from the databases with 80% and 40% of the entries differs from the generating structure in one link, and they are respectively $\exp(-980 + 985) = 148$ and $\exp(-993 + 995) = 7.4$ more likely than the generating structure. The estimates of the marginal probabilities are very similar to those obtained in the complete database, again showing that the consequence of a slightly different model has little effect on the reasoning process. The estimate of $-\log p(\mathcal{D}_i|\mathcal{M})$ is accurate until the database contains 40% of the original entries. The total run times make even clearer that the source of complexity is the search space and the performances of the method remain insensitive to the number of missing data. This result is not surprising when we realize that the computational cost of BC does not depend on the number of missing data. The number of missing data affects only the storage procedure described in (Ramoni and Sebastiani, 1997a) but its effect is limited by taking advantage of the local independencies of the BBN and by using discrimination trees to store the counters of observed data and to keep track of the possible completions.

| %   | Induced Model | $l(\mathcal{D}_i|\mathcal{M})$ | Time |
|-----|---------------|-------------------------------|------|
| 100 | $X_5 \leftarrow X_3 \quad\quad X_4$ <br> $\nwarrow \quad\quad \nearrow$ <br> $X_1 \longrightarrow X_2$ | 25024 | 183 |
| 80  | $X_5 \leftarrow X_3 \longrightarrow X_4$ <br> $\diagdown\!\!\!\diagup \quad \nearrow$ <br> $X_1 \longrightarrow X_2$ | 24673 | 191 |
| 60  | $X_5 \leftarrow X_3 \quad\quad X_4$ <br> $\diagdown\!\!\!\diagup \quad \nearrow$ <br> $X_1 \longrightarrow X_2$ | 24871 | 187 |
| 40  | $X_5 \leftarrow X_3 \longrightarrow X_4$ <br> $\diagdown\!\!\!\diagup \quad \nearrow$ <br> $X_1 \longrightarrow X_2$ | 24814 | 188 |
| 20  | $X_5 \leftarrow X_3 \quad\quad X_4$ <br> $\uparrow \quad \diagdown\!\!\!\diagup \quad \nearrow$ <br> $X_1 \longrightarrow X_2$ | 25112 | 185 |

Table 8: Models induced from the database generated from $\mathcal{M}_3$ for different percentages of available entries.

| %   | $X_1 = 1$ | $X_2 = 1$ | $X_3 = 1$ | $X_4 = 1$ | $X_5 = 1$ |
|-----|-----------|-----------|-----------|-----------|-----------|
| 100 | 0.20      | 0.70      | 0.39      | 0.30      | 0.52      |
| 80  | 0.20      | 0.71      | 0.38      | 0.29      | 0.53      |
| 60  | 0.21      | 0.70      | 0.39      | 0.29      | 0.53      |
| 40  | 0.21      | 0.70      | 0.39      | 0.30      | 0.53      |
| 20  | 0.21      | 0.69      | 0.40      | 0.31      | 0.54      |

Table 9: Marginal probabilities in the networks induced from the database generated from $\mathcal{M}_3$ for different percentages of available entries.

The models induced from the databases generated from $\mathcal{M}_3$ and $\mathcal{M}_4$ are given in Table 8 and 10, respectively. The initial order on the variables was in both cases $X_5 \prec X_4 \prec X_3 \prec X_2 \prec X_1$. The models induced from the complete database are equal to the generating structure for both $\mathcal{M}_3$ and $\mathcal{M}_\Delta$, and coherently the empty structure is induced when data are totally missing. Table 9 displays the marginal probabilities computed in the networks induced from the incomplete databases generated by $\mathcal{M}_3$.

As the number of entries available decreases, at most two extra dependencies are induced from the database. The only exception is the model induced from the database generated from $\mathcal{M}_4$ with 80% of the entries available. In this case, four extra dependencies are learned, and the Bayes factor of the induced model against the generating structure is $e^{13}$. However the conditional probabilities learned are only slightly different, so that the estimates of the marginal probabilities are extremely robust thus limiting the effect in the subsequent reasoning process. The estimates of $-\log p(\mathcal{D}_i|\mathcal{M})$ are again extremely accurate.

| % | Induced Model | | | $\hat{l}(\mathcal{D}_i|\mathcal{M})$ | Time |
|---|---|---|---|---|---|
| 100 | $X_5 \longleftarrow X_3 \quad X_4$ <br> $\nwarrow \qquad \nearrow$ <br> $X_1 \longrightarrow X_2$ | | | 40125 | 275 |
| 80 | $X_5 \longleftarrow X_3$ <br> $\uparrow \quad \not\times \quad \uparrow$ <br> $X_1 \longrightarrow X_2$ <br> $\searrow \quad \downarrow$ <br> $X_4$ | | | 39369 | 296 |
| 60 | $X_5 \longleftarrow X_3 \quad X_4$ <br> $\nwarrow \quad \uparrow \quad \nearrow$ <br> $X_1 \longrightarrow X_2$ | | | 42146 | 278 |
| 40 | $X_5 \longleftarrow X_3 \quad X_4$ <br> $\not\times \qquad \nearrow$ <br> $X_1 \longrightarrow X_2$ | | | 40132 | 289 |
| 20 | $X_5 \longleftarrow X_3 \quad X_4$ <br> $\nwarrow \quad \uparrow \quad \nearrow$ <br> $X_1 \longrightarrow X_2$ | | | 39952 | 285 |

Table 10: Models induced from the database generated from $\mathcal{M}_4$ for different percentages of available entries.

## 5   CONCLUSIONS

Missing data represent a challenge for learning methods because they may affect their use in real-world applications, where databases are often incomplete. Current methods to learn BBNs from incomplete databases rely on iterative methods, such as EM or Gibbs Sampling, to obtain an approximate estimate of the marginal likelihood of the database given a graphical model, a fundamental step in the process of extracting the graphical structure of a BBN from a database. This paper introduced a deterministic method able to provide this estimation, using BC, and to extract the graphical structure from an incomplete database. In this way, BC can be used to both induce the graphical structure and assess the conditional probabilities of a BBN from an incomplete database. Preliminary experimental evaluations show a significant robustness of this method and a remarkable independence of its execution time from the number of missing data.

### Acknowledgments

This research was partially supported by equipment grants from Apple Computers and Sun Microsystems.